\documentclass[10pt,twocolumn,letterpaper]{article}

\usepackage[pagenumbers]{cvpr} % To force page numbers, e.g. for an arXiv version
\usepackage{algorithm}
\usepackage{bm} 
\usepackage{appendix}
\usepackage{mathbbol}
\usepackage{xspace}
\usepackage{enumitem}
\usepackage{graphicx}  
\usepackage{caption} 
\usepackage{amsmath}
\usepackage{multirow}
\usepackage{colortbl}
\usepackage{pifont}
\usepackage{booktabs}
\usepackage{algorithm}
\usepackage{algpseudocode}
\usepackage[accsupp]{axessibility}  % Improves PDF readability for those with disabilities.

\definecolor{cvprblue}{rgb}{0.21,0.49,0.74}
\usepackage[pagebackref,breaklinks,colorlinks,allcolors=cvprblue]{hyperref}

\def\paperID{32769} % *** Enter the Paper ID here
\def\confName{CVPR}
\def\confYear{2026}

\title{Parameterized Prompt for Incremental Object Detection}

\author{
Zijia An$^{1,2}$ \quad
Boyu Diao$^{1,2}$\thanks{Corresponding Author.} \quad
Ruiqi Liu$^{1,2}$ \quad
Libo Huang$^{1,2}$ \quad
Chuanguang Yang$^{1,2}$ \quad
Fei Wang$^{1,2}$ \quad \\
Zhulin An$^{1,2}$ \quad
Yongjun Xu$^{1,2}$\\
$^{1}$State Key Laboratory of AI Safety, Institute of Computing Technology, Chinese Academy of Sciences\\
$^{2}$University of Chinese Academy of Sciences\\
{\tt\small
\{anzijia23p, diaoboyu2012, yangchuanguang, wangfei, anzhulin, xyj\}@ict.ac.cn} \\
{\tt\small
huanglibo@gmail.com \quad liuruiqi23@mails.ucas.ac.cn
\thanks{Accepted in Conference on Computer Vision and Pattern Recognition, 2026 (CVPR'26)}
}
}
\begin{document}
\maketitle
\begin{abstract}
Recent studies have demonstrated that incorporating trainable prompts into pretrained models enables effective incremental learning. However, the application of prompts in incremental object detection (IOD) remains underexplored. Our study reveals that existing prompt-pool-based approaches assume disjoint class sets across incremental tasks, which are unsuitable for IOD as they overlook the inherent co-occurrence phenomenon in detection. In co-occurring scenarios, unlabeled objects from previous tasks may appear in current task images, leading to confusion in prompts pool. In this paper, we hold that prompt structures should exhibit adaptive consolidation properties across tasks, with constrained updates to prevent confusion and catastrophic forgetting. Motivated by this, we introduce Parameterized Prompts for Incremental Object Detection (P$^2$IOD). Leveraging neural networks global evolution properties, P$^2$IOD employs networks as the parameterized prompts to adaptively consolidate knowledge across tasks. To constrain prompts structure updates, P$^2$IOD further engages a parameterized prompts fusion strategy. Extensive experiments on PASCAL VOC2007 and MS COCO datasets demonstrate that P$^2$IOD's effectiveness in IOD and achieves the state-of-the-art performance among existing baselines. Code is available at 
\url{https://github.com/EMLS-ICTCAS/P2IOD}.
\end{abstract}    
\section{Introduction}
\label{sec:intro}
In response to external changes, humans possess strong adaptability, allowing them to incrementally accumulate knowledge \cite{huang2025foundation, huang2024kfc}. Similarly, we expect object detection algorithms to learn in an incremental manner. Existing detection methods suffer from catastrophic forgetting \cite{mccloskey1989catastrophic} during incremental learning. This issue arises because current detection frameworks rely on predefined labeled datasets \cite{zou2023object}, implicitly assuming static data distributions. When learning from dynamic data distributions, these frameworks tend to forget previously learned knowledge \cite{wang2024comprehensive}, resulting in severe performance degradation.

\begin{figure}[tb]
    % \vspace{-5mm}
  \centering
  \includegraphics[width=\columnwidth]{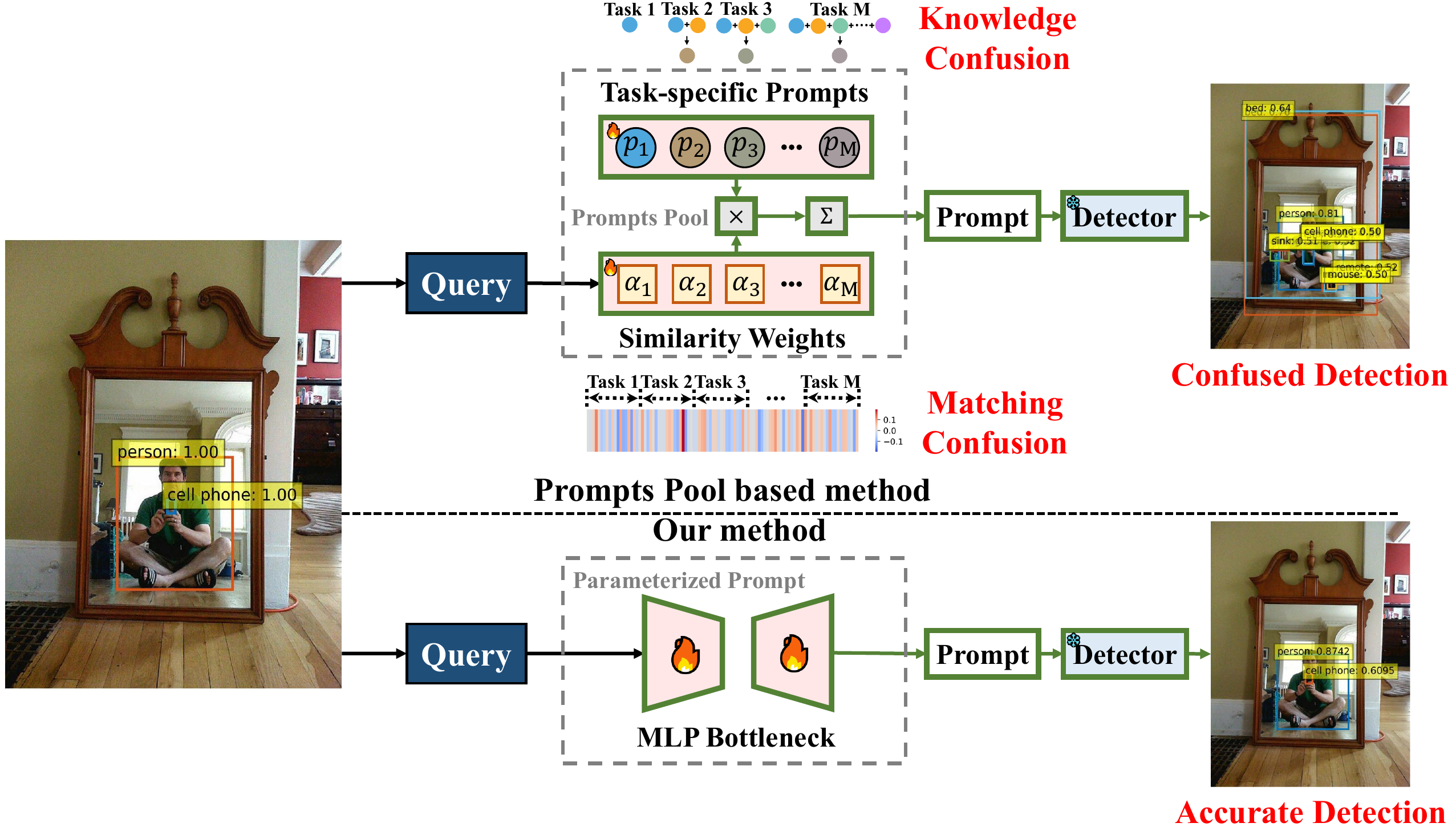}
  \caption{Existing prompts-pool-based methods store task-specific prompts learned from different tasks and match objects to their most relevant prompts based on similarity weights during inference. However, preserving knowledge in a task-isolated manner causes confusion in IOD. In contrast, our method redesigns the prompts pool as parameterized prompts to holistically preserve and update knowledge across tasks, thus mitigating confusion.}
      \vspace{-5mm}
  \label{fig: similarity_weight}
\end{figure}

To address this challenge, many incremental object detection (IOD) methods \cite{feng2022overcoming,liu2023continual, kim2024vlm, mo2024bridge, song2025learning} exploit the inherent co-occurrence phenomenon, where detection images typically contain both labeled objects from the current task and unlabeled objects from previous tasks. In such co-occurring scenarios, object distribution remains relatively static \cite{an2025ior}, providing latent knowledge to supplement previous tasks.  A key problem in IOD thus lies in effectively leveraging the static object distribution present in co-occurring scenarios. Recently, with the rise of pre-trained models \cite{cheng2026metagnsdformer}, prompting has emerged as a promising direction for incremental learning \cite{huang2025preprompt, li2025geometric}. Yet, its suitability for IOD’s co-occurring scenarios remains unexplored. Gaurav et al. \cite{bhatt2024preventing} first introduce prompting into IOD, adopting a well-established prompts pool from incremental classification. We observe that the prompts pool exhibits confusion when incorporating the knowledge of the static object distribution in co-occurring scenarios, leading to a negative impact on performance.

An ideal prompts pool stores task-specific prompts learned from different tasks and matches objects to its most relevant prompts based on similarity weight during inference \cite{wang2022learning}. However, when leveraging the static distribution of objects in co-occurring scenarios, the prompts pool encounters severe confusion, specifically manifesting as matching and task confusion. As shown in Fig.~\ref{fig: similarity_weight}, the former matching confusion refers to an object that cannot match the most relevant prompt. We visualize the similarity weight between an object and different prompts. It can be observed that since the object appears across all tasks, they exhibit high similarity with all task-specific prompts, making it impractical to match the most relevant prompt. On the other hand, the task confusion refers to task-specific prompts learning knowledge outside of its tasks. The unlabeled previous objects in co-occurring scenarios provide latent knowledge, causing the prompts learned for the current task to incorporate knowledge from all previous tasks, which undermines the clarity of the prompt's representation. We refer to the matching and task confusion introduced by the prompts pool in IOD as prompts pool confusion, which negatively affects IOD's performance.

To tackle the above problems, this paper proposes \textbf{P}arameterized \textbf{P}rompt for \textbf{I}ncremental \textbf{O}bject \textbf{D}etection (P$^2$IOD). We argue that preserving knowledge in a task-isolated manner leads to confusion when handling co-occurring scenarios in IOD. To overcome the confusion, we advocate that the structure for preserving prompt knowledge should exhibit an adaptive consolidation property, ensuring that knowledge is preserved holistically across tasks, while previous task knowledge can be dynamically updated in co-occurring scenarios. Building upon this, we further mitigate catastrophic forgetting by constraining updates to critical parameters within the prompt structure. Based on this insight, P$^2$IOD redesigns the prompts pool as a parameterized multi-layer perceptron (MLP), so as to leveraging the adaptive consolidation property inherent in neural networks, which naturally update learned knowledge in response to losses from co-occurring objects. We interpret the constraint on parameterized prompts as a form of model fusion, where the parameters of previous and current prompts are preserved or merged based on their importance and consistency, ensuring that the knowledge from each task is retained. In addition, we introduce pseudo-labeling to mine latent knowledge from co-occurring objects.

Our contributions can be summarized as follows.

(\romannumeral 1) This is the first work to investigate the prompts pool confusion caused by the co-occurrence phenomenon. We further advocate that prompt structures should exhibit an adaptive consolidation property and adopt constrained updates to prevent confusion and forgetting.

(\romannumeral 2) We propose P$^2$IOD, which redesigns the prompts pool as parameterized prompts and employs parameterized prompt fusion to constrain parameter updates.

(\romannumeral 3) Extensive experiments on PASCAL VOC2007 and MS COCO datasets demonstrate the effectiveness of the proposed method in IOD, achieving state-of-the-art performance in existing baselines.
\begin{figure*}[tb]
    % \vspace{-7mm}
  \centering
  \includegraphics[width=2.0\columnwidth]{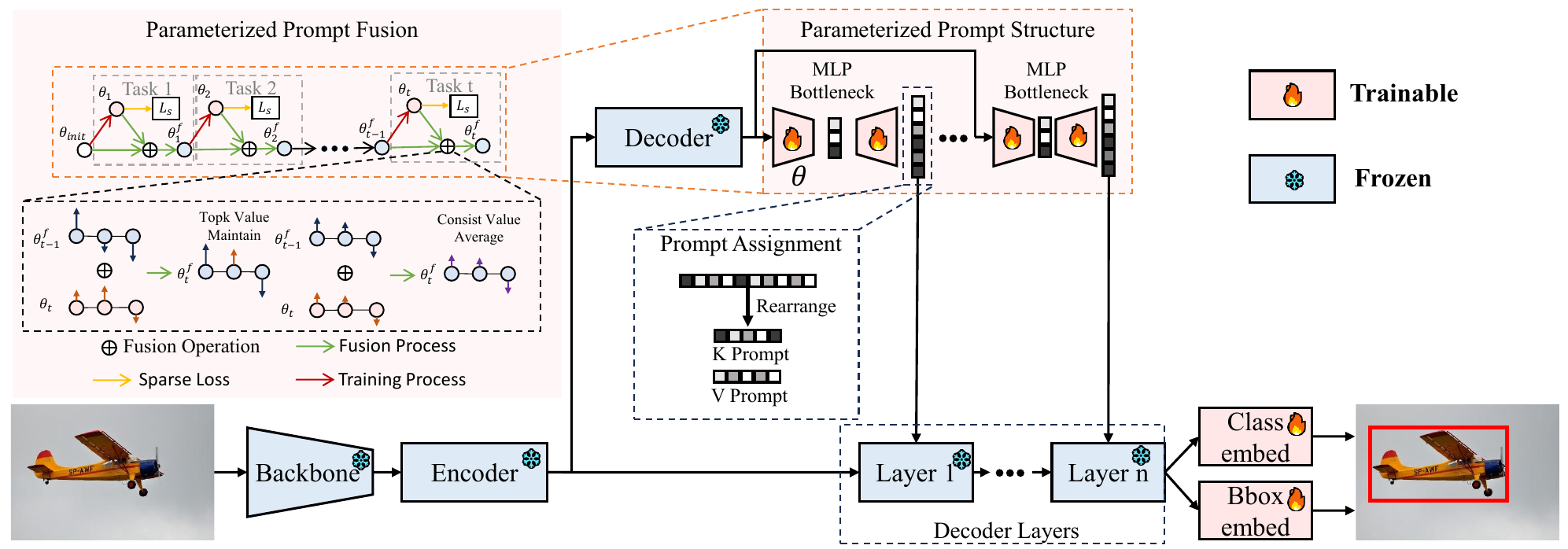}
      % \vspace{-2mm}
  \caption{The overall framework of P$^2$IOD. To address the issue of prompts pool confusion, P$^2$IOD redesigns the prompts pool as a parameterized prompt structure consisting of multi-layer perceptron (MLP) bottlenecks. To further alleviate catastrophic forgetting, P$^2$IOD proposes a parameterized prompt fusion strategy, which adds an additional fusion process after each incremental training process.}
      \vspace{-5mm}
  \label{fig: Overall structure}
\end{figure*}

\section{Related Work}
\label{sec: rela}

\subsection{Incremental Learning}
In recent years, the strong generalization ability of pre-trained models (PTM) injects new vitality into incremental learning \cite{zhou2024continual}. A promising approach is to freeze the PTM's parameters and add learnable lightweight prompts to adjust the PTM \cite{wang2022learning, wang2022dualprompt, smith2023coda, jung2023generating}. However, the learnable prompts also face the challenging issue of catastrophic forgetting. L2P \cite{wang2022learning} and DualPrompt \cite{wang2022dualprompt} design a prompts pool to store task-specific prompts trained under different tasks. During inference, the top-K most relevant prompts are selected through an instance-wise query mechanism, thereby alleviating the catastrophic forgetting caused by updating prompts. CodaPrompt \cite{smith2023coda} replaces the top-K selection criterion with a more natural selection mechanism, using a learnable linear combination to determine the contribution of the prompts. DAP \cite{jung2023generating} uses an MLP to generate finer-grained prompts for each instance and utilizes a prompts pool to store conditional input embeddings that supplement task-specific information. The above prompt-based methods are discussed in the context of incremental image classification tasks and show remarkable results, but their applicability in more complex incremental object detection tasks is still not fully established.

\subsection{Incremental Object Detection}

The distinction between incremental object detection and other incremental tasks lies in the co-occurrence phenomenon inherent in detection scenarios. Co-occurring scenarios contain numerous unlabeled objects from previous tasks that can supplement previous task knowledge. Knowledge distillation \cite{hinton2015distilling, huang2024etag, liu2025low, yu2025merlin, wu2025navigating} provides a flexible way to mine previous task knowledge. Such approaches \cite{shmelkov2017incremental,feng2022overcoming,cermelli2022modeling, yang2022rd, yang2022multi} employ the original detector to regularize the outputs and intermediate features of the incremental detector, thereby facilitating the transfer of knowledge in training data from the original to the incremental detector. As this knowledge inherently contains information about unlabeled objects, it enables the implicit mining of unlabeled objects within the co-occurring scenarios. Furthermore, some methods \cite{kim2024vlm, bhatt2024preventing, yang2023pseudo} explicitly mine unlabeled objects through pseudo-labeling. These methods use the original detector to label the objects in the training data and subsequently filter out incorrect labels based on specific criteria. The static distribution labels obtained through the pseudo-labeling method allow the detectors to be immune to catastrophic forgetting. The above methods exhibit strong performance in co-occurring scenarios. 

With the rise of PTM, the prompting of the PTM has become a promising direction for incremental learning. Prompting of the PTM stores prompts as an additional memory module, allowing the PTM to learn and retain relevant information from each incremental task. Gaurav \cite{bhatt2024preventing} constructs a prompts pool to store the prompts learned from different tasks and matches the most relevant prompt during inference. However, we discover that introducing a prompts pool faces severe prompts pool confusion in co-occurring scenarios. Therefore, effectively incorporating prompts into PTM still requires further research in IOD.

\section{Preliminaries}
\label{sec: pre}

\subsection{Object Detection Baseline}
We introduce parameterized prompts on the transformer-based Deformable-DETR \cite{zhu2020deformable} and Co-DETR \cite{zong2023detrs} to validate our motivation. In transformer-based object detection frameworks, there exist two types of attention mechanisms. The first is the multi-head attention mechanism \cite{vaswani2017attention} in Transformers. Given a query element and a set of key elements, the multi-head attention module obtains attention weights based on the similarity between the query-key pairs and adaptively aggregates important features according to the attention weights. To enable the model to focus on content from different representational subspaces and different positions, the multi-head attention mechanism combines the outputs of multiple attention heads with different learnable weights.  Let $q \in {\Omega _q}$ indexes a query element with representation feature ${z_q} \in {\mathbb{R}^C}$, and $k \in {\Omega _k}$ indexes a key element with representation feature ${x_k} \in {\mathbb{R}^C}$, where $C$ is the feature dimension, ${\Omega _q}$ and ${\Omega _k}$ specify the query and key elements, respectively. The attention weights ${A_{mqk}}$ are calculated by 
\begin{equation}
{A_{mqk}} = {\rm{softmax}}\left( {\frac{{z_q^TU_m^T{V_m}{x_k}}}{{\sqrt {{C_v}} }}} \right),
\label{A_{mqk}}
\end{equation}
where ${U_m} \in {\mathbb{R}^{{C_v} \times C}}$ and ${V_m} \in {\mathbb{R}^{{C_v} \times C}}$ are learnable weights of $q$ and $k$. The process of calculating the aggregated features in the multi-head attention mechanism can be represented by
\begin{equation}
{\text{MHA}}\left( {{z_q},x} \right) = \sum\limits_{m = 1}^M {{W_m}\left[ {\sum\limits_{k \in {\Omega _k}} {{A_{mqk}} \cdot {{W'}_m}{x_k}} } \right]},
\label{MultiHeadAttn}
\end{equation}
where $m$ indexes the attention heads, ${{W'}_m} \in {\mathbb{R}^{{C_v} \times C}}$ and ${W_m} \in {\mathbb{R}^{C \times {C_v}}}$ are also learnable weights (${C_v} = {C \mathord{\left/ {\vphantom {C M}} \right. \kern-\nulldelimiterspace} M}$). Moreover, to disambiguate different spatial positions, the representation features ${z_q}$ and ${x_k}$ are usually introduced with positional embeddings.

The second attention mechanism is the deformable attention mechanism \cite{zhu2020deformable}. It not only preserves the spatial structure of the feature map but also helps the detector accelerate convergence and reduce computational complexity. Given an input feature map $x \in {\mathbb{R}^{C \times H \times W}}$, let $q$ index a query element with content feature ${z_q}$ and a 2-d reference point $p_q$, the deformable attention feature is calculated by
\begin{equation}
{\text{DA}}\left( {{z_q},{p_q},x} \right) = \sum\limits_{m = 1}^M {{W_m}\left[ {\sum\limits_{k = 1}^K {{A_{mqk}} \cdot {{W'}_m}x\left( {{p_q} + \Delta {p_{mqk}}} \right)} } \right]},
\label{DeformAttn}
\end{equation}
where $m$ indexes the attention head, $k$ indexes the sampled keys, and $K$ is the total sampled key number ($K \ll HW$). ${\Delta {p_{mqk}}}$ and ${{A_{mqk}}}$ denote the sampling offset and attention weight of ${k^{{\text{th}}}}$ sampling point in the ${m^{{\text{th}}}}$ attention head, respectively. Both ${\Delta {p_{mqk}}}$ and ${{A_{mqk}}}$ are obtained via linear projection over the query feature ${z_q}$.

To the best of our knowledge, there is currently no research integrating prompts into deformable attention mechanisms. The difficulty lies in the fact that deformable attention is a spatially local attention structure, while prompt interaction requires a global attention structure, making their integration challenging. Therefore, we only introduce prompts in the multi-head attention mechanism, which is used for object query interaction in the decoder.

\subsection{Pseudo Labeling for Mining Potential Objects}
In IOD, unlabeled previous task objects may appear in the background of current task images. Properly mining these previous task objects can significantly reduce forgetting, while treating these objects as background can lead to more severe forgetting. To make a fair comparison, we introduce the same pseudo-labeling method as MD-DETR \cite{bhatt2024preventing} to mine the knowledge of unlabeled objects in the background.

In $T_t$, we employ the detector trained on $T_{t-1}$ to infer on each training sample, obtaining predictions: ${{\hat y}_i} = \left\{ {{{\hat s}_i},{{\hat b}_i}} \right\}$. Here, ${{{\hat s}_i}}$ represents the score for the highest-scoring category, and ${{{\hat b}_i}}$ provides the bounding box coordinates for this prediction. Pseudo-labeling mechanism \cite{dong2022open, gupta2022ow, liu2023continual} sets a threshold $ \tau $ to filter predictions with ${{{\hat s}_i}}$ higher than this threshold as pseudo label ${{\tilde y}_i} = \left\{ {{{\tilde c}_i},{{\tilde b}_i}} \right\}$. The threshold $ \tau $ ensures that only the most reliable predictions are used when generating pseudo labels. Here, ${{{\tilde c}_i}}$ represents the pseudo label's category name, ${{{\tilde b}_i}}$ is the bounding box coordinates for this pseudo label. Pseudo labels incorporate the knowledge from previous tasks $\left\{ {{T_1} \ldots {T_{t - 1}}} \right\}$, effectively alleviating the detector's forgetting.
\section{METHODS}
\label{sec: methods}

\subsection{Overview}
We propose P$^2$IOD to alleviate the confusion in prompt-pool-based IOD methods when learning co-occurring object knowledge. Fig.~\ref{fig: Overall structure} illustrates the complete framework. We hold that prompt structures should exhibit the ability to adaptively consolidate knowledge across tasks while constraining updates to prevent catastrophic forgetting. Therefore, P$^2$IOD redesigns the prompts pool into parameterized prompts, leveraging neural networks' inherent adaptive consolidation to naturally update learned knowledge in response to losses from co-occurring objects. The parameterized prompts are implemented as multi-layer perceptron (MLP) bottlenecks composed of feedforward networks and are integrated into different decoder layers to enhance prompt diversity. Considering that IOD algorithms are often deployed on resource-constrained scenarios such as edge devices \cite{liu2024continual, liu2025gensor}, P$^2$IOD proposes a parameterized prompt fusion strategy to constrain updates to the prompt structure with low computational overhead. During incremental training, only the parameters of class embeddings, bounding box embeddings, and the parametrized prompt structure ($\theta $) are trainable, while all other parameters ($\theta ^ * $) remain frozen to prevent knowledge forgetting. Fig.~\ref{fig: Overall structure} illustrates the complete framework.

\subsection{Parameterized Prompt Structure}
We hold that the prompt structure should adaptively consolidate the potential knowledge that emerges in the co-occurring scenarios. To achieve this, we design the prompt structure as an MLP bottleneck composed of FNN layers rather than the prompts pool. This parameterized prompt structure encodes prompt-related knowledge into the neural network weight space and generates instance-specific prompts. 

We follow the method in \cite{bhatt2024preventing} by employing the frozen pre-trained detector as a query function to extract queries, which are then utilized as inputs to the parameterized prompt. Given an input instance $x$, a set of proposals $P \in {\mathbb{R}^{N \times D}}$ is generated through a single pass of $P = {\theta ^ * }\left( x \right)$, where $N$ is the number of proposals and $D$ is the embedding dimension of each proposal. $P$ contains both object and background information related to the instance $x$, which is preliminarily extracted by the frozen pre-trained detector. However, the number of proposals in $P$ is too large to be directly used as query features, requiring compression. Unlike the approach in \cite{bhatt2024preventing}, where only object-related proposals are compressed, we find that jointly compressing both object and background proposals in P$^2$IOD is more effective. We believe this is because including background knowledge in the prompts helps the detector better distinguish between foreground and background. In contrast, \cite{bhatt2024preventing} shows that adding background query features degrades its performance. This degradation stems from the knowledge retention and matching mechanism of the prompts pool, which limits the acquisition of background knowledge, indicating that the prompts pool structure is not suitable for IOD. Detailed analysis and experimental comparisons can be found in Appendix B.4. The entire query function $Q$ can be represented as follows:
\begin{equation}
Q\left( {x,{\theta ^ * }} \right) = \frac{1}{N}\sum\limits_{n = 1}^N {{{\left\{ {{\theta ^ * }\left( x \right)} \right\}}_n}} ,
\label{query function}
\end{equation}
where ${{{\left\{ {{\theta ^ * }\left( x \right)} \right\}}_n}}$ is the ${n^{{\text{th}}}}$ proposal.

We take the $Q\left( {x,{\theta ^ * }} \right)$ as the input to the parameterized prompt, which outputs the prompts $p \in {\mathbb{R}^{{L_p} \times D}}$. ${L_p}$ represents the length of prompts. The parameterized prompt is an MLP bottleneck composed of two FNN layers, which can effectively remove redundant information in the query through linear dimensionality reduction. The entire process can be represented as:
\begin{equation}
p = {\text{ReLU}}\left( {Q\left( {x,{\theta ^ * }} \right) \cdot {W^{\left( 1 \right)}}} \right) \cdot {W^{\left( 2 \right)}},
\label{MLP}
\end{equation}
where ${W^{\left( 1 \right)}} \in {\mathbb{R}^{D \times d}}$ represents an FNN layer for dimensionality reduction, in which $d$ is the bottleneck dimension; ${W^{\left( 2 \right)}} \in {\mathbb{R}^{d \times \hat D}}$ is an FNN layer with upper-projection parameters, where $\hat D = D \times {L_p}$; RELU is non-linear activation in between.

The $p \in {\mathbb{R}^{{L_p} \times D}}$ are integrated into the decoder's multi-head self-attention layers. The process can be expressed as follows:
\begin{equation}
{\text{MH}}{{\text{A}}_p}\left( {{q_o},p} \right) = \sum\limits_{m = 1}^M {{W_m}\left[ {\sum\limits_k {{A_{mqk}} \cdot {\left[ {{W'}_mq_o}:{p_v}\right]} } } \right]},
\label{MHA with prompt}
\end{equation}
\begin{equation}
{A_{mqk}} = {\text{softmax}}\left( {\frac{{q_o^TU_m^T{\left[ {V_mq_o}:{p_k}\right]} }}{{\sqrt {{C_v}} }}} \right),
\label{A_mqk with prompt}
\end{equation}
where ${{q_o}}$ is the objects queries \cite{zhu2020deformable}, $\left[ {x:y} \right]$ represents the concatenate operation. Following \cite{wang2022dualprompt}, we assign $p \in {\mathbb{R}^{{L_p} \times D}}$ into ${p_k} \in {\mathbb{R}^{\frac{{{L_p}}}{2} \times D}}$ and ${p_v} \in {\mathbb{R}^{\frac{{{L_p}}}{2} \times D}}$, and concatenate them to ${{W'}_m}{q_o}$ and ${V_m}{q_o}$ respectively, while keeping ${q_o^TU_m^T}$ unchanged. This manner ensures that the input and output sequence lengths remain the same before and after integrating prompts. To increase prompt diversity, we introduce independent parameterized prompts into each decoder layer of the frozen pre-trained detector.

\subsection{Parameterized Prompt Fusion for Incremental Learning}
The parameterized prompt also faces catastrophic forgetting during incremental learning. To address the forgetting, we introduce model fusion after each incremental training process. During the fusion process, we aim to retain the important parameters of each task and average the consistent parameters across tasks. We also introduce a sparse loss to concentrate the knowledge of each task in different parameter subsets to facilitate the model fusion.

\noindent\textbf{Model fusion.} For a sequence of incremental tasks $\left\{ {{T_1} \ldots {T_t}} \right\}$, we add a fusion process after the training process in $\left\{ {{T_2} \ldots {T_t}} \right\}$ to fuse the parameterized prompt of the current task with those of the previous task. We denote the parameterized prompt obtained from training as $\theta _t$ and those obtained from fusion as $\theta _t^f$. For $T_t$ ($t \geqslant 2$), the parameterized prompt used for testing is $\theta _t^f$, which is obtained by fusing $\theta _t$ and $\theta _{t - 1}^f$ (when $t=2$, we fuse $\theta _2$ and $\theta _1$). 

We fuse $\theta _t$ and $\theta _{t - 1}^f$ based on the degree of parameter variation. To describe the variation of parameterized prompt between current and previous tasks ($\theta_t$ and $\theta_{t-1}^f$), we compute the task vector ${\bm{v}_t} = {\theta _t} - \theta _{t - 1}^f$. The task vector ${\bm{v}_t}$ simultaneously conveys the parameter variation's magnitude and direction. Inspired by \cite{yadav2023ties}, we decompose the task vector $\bm{v}_t$ into a magnitude vector $\mu _t$ ($\mu _t = \left| {\bm{v}_t} \right|$) and a sign vector $\gamma _t$ ($\gamma _t = {\text{sgn}}\left( {\bm{v}_t} \right)$, taking values in $\pm 1$) as $\bm{v}_t = \gamma _t \odot \mu _t$, where $ \odot $ is the element-wise product. We also describe the overall variation of parameterized prompt in previous tasks by computing the task vector $\bm{v}_{t - 1}^{f} = \theta _{t - 1}^f - {\theta _{init}}$, where $\theta _{init}$ denotes the initialized parameterized prompt.

During the $T_t$ fusion process, we preserve critical parameters guided by $\mu_t$ and $\mu_{t - 1}^{f}$, and average consistent parameters based on $\gamma_t$ and $\gamma_{t - 1}^{f}$. To preserve critical parameters, we first sort the values in $\mu_{t - 1}^{f}$ and select the ${\text{top-}}k$\% indices, denoted as $\mathcal{I}_{t - 1}^{f}$. The corresponding parameter in $\theta _{t - 1}^f$ are preserved with priority at the indices in $\mathcal{I}_{t - 1}^{f}$. Next, we sort $\mu_t$ and identify the ${\text{top-}}l$\% indices, denoted as $\mathcal{I}_t$. The parameter in ${\theta_t}$ are preserved at the indices in $\mathcal{I}_t$, excluding any overlap with the indices in $\mathcal{I}_{t-1}^{f}$. To average consistent parameters, we locate positions where $\gamma_t = \gamma_{t - 1}^{f}$, indicating directional consistency. At these positions, excluding those already reserved for preservation (i.e., $\mathcal{I}_{t-1}^{f} \cup \mathcal{I}_t$), we take the average of $\theta_t$ and $\theta_{t-1}^{f}$. Finally, all remaining undecided parameters are assigned the corresponding values from $\theta_{t-1}^{f}$. The overall fusion process can be formally expressed as follows:
\begin{equation}
\theta _t^f[i] = \left\{ 
\begin{array}{ll}
  \theta _{t - 1}^f[i], & i \in \mathcal{I}_{t - 1}^{f} \\ 
  \theta _t[i], & i \in \mathcal{I}_t \setminus \mathcal{I}_{t - 1}^{f} \\ 
  \frac{1}{2}(\theta _t[i] + \theta _{t - 1}^f[i]), & \gamma_t[i] = \gamma_{t - 1}^{f}[i],\; i \notin \mathcal{I}_{t - 1}^{f} \cup \mathcal{I}_t \\ 
  \theta _{t - 1}^f[i], & \text{otherwise}
\end{array} 
\right.
\label{Model Fusion}
\end{equation}
where $i$ denotes the $i$-th parameter in the parameterized prompts. The pseudo-code for parameterized prompt fusion is outlined in Appendix C.1.

\noindent\textbf{Sparse Loss.} In model fusion, we retain the important parameters of both current and previous tasks to maintain the learned knowledge. However, in practice, the learned parameters exhibit redundancy, making it difficult to identify parameter importance. We expect the model to learn sparse parameters to concentrate critical knowledge in a small subset of parameters. Therefore, we introduce an additional ${L_1}$ loss as a sparse loss ${L_s}$, defined as:
\begin{equation}
{L_s} = \lambda \sum\limits_j {\left| {\theta _j} \right|} ,
\label{integrate prompt}
\end{equation}
where $\lambda$ controlls the sparsity level, and ${\theta _j}$ refers to the parameterized prompts in the ${j^{{\text{th}}}}$ decoder layer.

\begin{table*}[ht]
\centering
\caption{Average precision ($A{P_{50}}$, \%) is compared on the PASCAL VOC2007 dataset under single-step settings of 19+1, 15+1, 10+10, and 5+15. We add the superscript $^ *$ to the accuracy that may be overestimated.  The reasons for the overestimation are detailed in \ref{sec:Experimental Settings}.}
\label{tab: VOC One-step Comparisons}
\setlength{\tabcolsep}{9pt}
\begin{tabular}{l||ccc|ccc|ccc}
\hline
\multirow{2}{*}{\textbf{Method}} & \multicolumn{3}{c|}{\textbf{19+1}} & \multicolumn{3}{c|}{\textbf{15+5}} & \multicolumn{3}{c}{\textbf{10+10}} \\
\cline{2-10} 
& \textbf{1-19} & \textbf{20} & \textbf{1-20} & \textbf{1-15} & \textbf{16-20} & \textbf{1-20} &  \textbf{1-10} & \textbf{11-20} & \textbf{1-20}\\
\hline
\hline
% ORE \cite{joseph2021towards} & 69.4 & 60.1 & 68.9 & 71.8 & 58.7 & 68.5 & 60.4 & 68.8 & 64.6 \\
OW-DETR \cite{gupta2022ow} & 70.2 & 62.0 & 69.8 & 72.2 & 59.8 & 69.1 & 63.5 & 67.9 & 65.7\\
% ILOD-Meta & 70.9 & 57.6 & 70.2 & 71.7 & 55.9 & 67.8 & 68.4 & 64.3 & 66.3\\
ABR \cite{liu2023augmented} & 71.0 & 69.7 & 70.9 & 73.0 & 65.1 & 71.0 & 71.2 & 72.8 & 72.0 \\
% \hline
Faster ILOD \cite{peng2020faster} & 68.9 & 61.1 & 68.5 & 71.6 & 56.9 & 67.9 & 69.8 & 54.5 & 62.1 \\
% PPAS & 70.5 & 53.0 & 69.2 & - & - & - & 63.5 & 60.0 & 61.8 \\
% MVC & 70.2 & 60.6 & 69.7 & 69.4 & 57.9 & 66.5 & 66.2 & 66.0 & 66.1 \\
PROB \cite{zohar2023prob} &73.9 & 48.5 & 72.6 & 73.5 & 60.8 & 70.1 & 66.0 & 67.2 & 66.5\\
PseudoRM \cite{yang2023pseudo} & 72.9 & 67.3 & 72.6 & 73.4 & 60.9 & 70.3 & 69.1 & 68.6 & 68.9 \\
% MMA \cite{cermelli2022modeling} & 71.1 & 63.4 & 70.7 & 73.0 & 60.5 & 69.9 & 69.3 & 63.9 & 66.6 \\
BPF \cite{mo2024bridge} & 74.5 & 65.3 & 74.1 & 75.9 & 63.0 & 72.7 & 71.7 & 74.0 & 72.9 \\
% \hline
VLM-PL \cite{kim2024vlm} & 73.7$^ * $ & 89.3$^ * $ & 73.6 & 73.9$^ * $ & 82.4$^ * $ & 72.4 & 80.3$^ * $ & 76.3$^ * $ & 78.3 \\
\hline
% MD-DETR \cite{bhatt2024preventing} & 76.8$^ * $ & 67.2$^ * $ & 76.1 & 77.4$^ * $ & 69.4$^ * $ & 76.7 & 73.1$^ * $ & 77.5$^ * $ & 73.2 \\
% P$^2$IOD (our) & 78.5& 62.6 & \textbf{77.7} & 83.3 & 66.9 & \textbf{79.2} & 81.9 & 80.5 & \textbf{81.2} \\
% w/o prompt (MS COCO) & 75.2 & 55.3 & 74.2 & 81.5 & 53.6 & 74.5 & 80.6 & 76.4 & 78.5 \\
MD-DETR (MS COCO) \cite{bhatt2024preventing} & 76.8$^ * $ & 67.2$^ * $ & 76.1 & 77.4$^ * $ & 69.4$^ * $ & 76.7 & 73.1$^ * $ & 77.5$^ * $ & 73.2 \\
P$^2$IOD (MS COCO) & 78.5& 62.6 & \textbf{77.7} & 83.3 & 66.9 & \textbf{79.2} & 82.0 & 80.4 & \textbf{81.2}\\
\hline
% w/o prompt (Objects365) &  &  &  &  &  &  &  &  &  \\
MD-DETR (Objects365) \cite{bhatt2024preventing} & 89.4 & 68.7 & 88.3 & 86.1 & 84.7 & 85.8 & 81.8 & 87.3 & 84.6 \\
P$^2$IOD (Objects365) & 89.7 & 77.5 & \textbf{89.1} & 91.2 & 85.2 & \textbf{89.7} & 88.4 & 91.1 & \textbf{89.8}\\
\hline
\end{tabular}
\end{table*}

\section{EXPERIMENTS}
\label{sec: Experiments}

\subsection{Experimental Settings}
\label{sec:Experimental Settings}

\textbf{Datasets.} We evaluate our proposed method on PASCAL VOC2007
\cite{everingham2010pascal} and MS COCO \cite{lin2014microsoft}. The PASCAL VOC2007 contains 20 diverse object classes, including 9,963 images, split into 5,011 for training and 4,952 for testing. The MS COCO, with its 80 object classes spread across 118,000 training images and 5,000 evaluation images, makes it a more challenging benchmark.

\noindent\textbf{Eval metrics.} Followed by \cite{bhatt2024preventing}, we use the mean average precision at an IOU threshold of 0.5 ($A{P_{50}}$, \%) as the metric. For PASCAL VOC2007, following previous works \cite{kim2024vlm}, we provide the $A{P_{50}}$ of the current task classes and the previous task classes to better reflect the method's stability and plasticity. There are two evaluation methods for obtaining task precision: validating on the entire test set versus using task-specific test subsets. The second method yields higher precision for the same detector. We adopt the first method and add superscript $^ * $ to results from the second method to ensure fair comparison. For MS COCO, following previous works \cite{kim2024sddgr}, we provide the $A{P_{50}}$ of all learned classes after learning at each task.

\noindent\textbf{Implementation details.} 
We implement our proposed method based on Deformable-DETR \cite{zhu2020deformable} pre-trained on the MS COCO dataset and Co-DETR \cite{zong2023detrs} pre-trained on the Objects365 dataset \cite{shao2019objects365}, both obtained from HuggingFace. The large-scale Objects365 dataset contains 365 categories and over 600,000 images, making it a suitable pre-training source for incremental learning experiments on MS COCO dataset. Furthermore, since the official MD-DETR implementation is only available on Deformable-DETR, we port MD-DETR \cite{bhatt2024preventing} on Co-DETR for a fair comparison.

\begin{table*}[ht]
\centering
\caption{Average precision ($A{P_{50}}$, \%) is compared on the MS COCO dataset under multi-step settings of 40+20+20 and 40+10+10+10+10.}
\label{tab: COCO Multi-step Comparisons}
\setlength{\tabcolsep}{14pt}
\vspace{-2mm}
\begin{tabular}{l||c|cc|cccc}
\hline
\multirow{2}{*}{\textbf{Method}} & \multirow{2}{*}{$\mathcal{T}_1$ (1-40) } & \multicolumn{2}{c|}{\textbf{40+20+20}} & \multicolumn{4}{c}{\textbf{40+10+10+10+10}}\\
\cline{3-8} 
% & & $\mathcal{T}_2$ (41-60) & $\mathcal{T}_3$ (61-80) & $\mathcal{T}_2$ (41-50) & $\mathcal{T}_3$ (51-60) & $\mathcal{T}_4$ (61-70) & $\mathcal{T}_5$ (71-80)  \\
& & $\mathcal{T}_2$ & $\mathcal{T}_3$ & $\mathcal{T}_2$ & $\mathcal{T}_3$ & $\mathcal{T}_4$ & $\mathcal{T}_5$  \\
\hline
\hline
% SID \cite{peng2021sid} & 63.7 & 51.6 & 36.6 &  52.1 &  38.0 &  23.0 &  23.3 \\
ERD \cite{feng2022overcoming} & 63.7 & 54.5 & 48.6 & 53.9 & 46.7 & 39.9 & 31.8  \\
CL-DETR \cite{liu2023continual} & 63.7 & 58.3 & 54.1 & 54.4 & 50.2 & 45.6 & 38.2 \\
DyQ-DETR \cite{zhang2025dynamic} & 63.7 & 57.0 & 55.7 & 55.9 & 53.8 & 50.8 & 49.8\\
SDDGR \cite{kim2024sddgr} & 68.6 & 62.6 & 59.5 &  62.8 & 60.2 & 59.0 & 54.7 \\
LEA \cite{song2025learning} & 75.3 & 62.0 & 57.5 & 66.3 & 61.7 & 59.7 & 56.5 \\
GCD \cite{wang2025gcd} & - & - & 60.4 &  - & - & - & 55.1 \\
\hline
% w/o prompt (Objects365) & 78.2 & 62.9 & 36.4 &  &  &  & \\
MD-DETR (Objects365) \cite{bhatt2024preventing}& 79.0 & 69.4 & 60.3 & 68.1 & 61.7 & 53.7 & 49.4\\
P$^2$IOD (Objects365) & 79.6 & 71.3 & \textbf{68.8} & 74.1 & 70.9 & 69.3 & \textbf{64.8}\\
\hline
\end{tabular}
\end{table*}

\begin{table*}[ht]
\centering
\caption{Ablation study results ($A{P_{50}}$, \%) for component's contribution
evaluated on PASCAL VOC2007 in 5+5+5+5 setting.}
\label{tab: ablation study}
\setlength{\tabcolsep}{15pt}
\vspace{-2mm}
\begin{tabular}{c|c|c|c|c|ccc}
\hline
\multirow{2}{*}{Methods} & Pseudo & Parameterized & Model & Sparse & \multicolumn{3}{c}{\textbf{5+5+5+5}} \\
\cline{6-8} 
 & Labeling & Prompt Structure & Fusion & Loss & \textbf{1-5} & \textbf{6-20} & \textbf{1-20} \\
\hline
\hline
(a) & & & & & 73.3 & 65.4 & 67.4 \\
(b) & \checkmark & & & & 73.3 & 64.6 &66.8 \\
(c) & \checkmark & \checkmark & & & 70.7 & 76.6 & 75.1 \\
(d) & \checkmark & \checkmark & \checkmark & & 73.1 & 76.0 & 75.3 \\
(e) &  & \checkmark & \checkmark & \checkmark & 67.0 & 73.0 & 71.5 \\
(f) & \checkmark & \checkmark & \checkmark & \checkmark & \textbf{74.0} & \textbf{77.2} & \textbf{76.4} \\
\hline
\end{tabular}
\vspace{-3mm}
\end{table*}

\subsection{Comparison}

{
\setlength{\tabcolsep}{1mm}
\begin{table}[t]
\centering
\caption{Average precision ($A{P_{50}}$, \%) is compared on the PASCAL VOC2007 dataset under multi-step settings of 10+5+5 and 5+5+5+5. We add the superscript $^ *$ to the accuracy may be overestimated. The reasons for the overestimation are detailed in \ref{sec:Experimental Settings}.}
\label{tab: VOC Multi-step Comparisons}
\footnotesize
\vspace{-2mm}
\begin{tabular}{l||ccc|ccc}
\hline
\multirow{2}{*}{\textbf{Method}} & \multicolumn{3}{c|}{\textbf{10+5+5}} & \multicolumn{3}{c}{\textbf{5+5+5+5}}\\
\cline{2-7} 
% & \textbf{1-10 ($T_1$)} & \textbf{10-20 ($T_2+T_3$)}  & \textbf{1-20} &\textbf{1-5 ($T_1$)} & \textbf{6-20 ($T_2+T_3+T_4$)} & \textbf{1-20}\\
& \textbf{1-10} & \textbf{10-20}  & \textbf{1-20} &\textbf{1-5} & \textbf{6-20} & \textbf{1-20}\\
\hline
\hline
ABR \cite{liu2023augmented} & 68.7 & 67.1 & 67.9 & 64.7 & 56.4 & 58.4 \\
% \hline
% ILOD & 67.2 & 59.4 & 63.3 & 58.5 & 15.6 & 26.3  \\
Faster ILOD \cite{peng2020faster} & 68.3 & 57.9 & 63.1 & 55.7 & 16.0 & 25.9\\
MMA \cite{cermelli2022modeling} & 67.4 & 60.5 & 64.0 & 62.3 & 31.2 & 38.9\\
BPF \cite{mo2024bridge} & 69.1 & 68.2 & 68.7 & 60.6 & 63.1 & 62.5\\
% \hline
VLM-PL \cite{kim2024vlm} & 67.9$^ * $ & 67.9$^ * $ & 67.9 & 64.5$^ * $ & 68.4$^ * $ & 65.5\\
\hline
% MD-DETR \cite{bhatt2024preventing} & 68.5 & 60.3 & 60.7 & 55.2 & 63.6 & 61.5\\
% P$^2$IOD (our) & 81.3 & 74.2 & \textbf{77.8} & 73.7 & 77.2 & \textbf{76.3} \\
% w/o prompt (MS COCO) & 80.6 & 64.0 & 72.3 & 73.3 & 65.4 & 67.4\\
MD-DETR (MS COCO) \cite{bhatt2024preventing}& 68.5 & 60.3 & 60.7 & 55.2 & 63.6 & 61.5\\
P$^2$IOD (MS COCO)& 81.3 & 74.2 & \textbf{77.8} & 74.0 & 77.2 & \textbf{76.4} \\
\hline
% w/o prompt (Objects365) &  &  &  &  &  & \\
MD-DETR (Objects365) \cite{bhatt2024preventing} & 80.1 & 87.5 & 83.8 & 60.9 & 80.7 & 75.8\\
P$^2$IOD (Objects365)& 89.1 & 89.1 & \textbf{89.1} & 86.4 & 87.4 & \textbf{87.1} \\
\hline
\end{tabular}
\vspace{-5mm}
\end{table}

}

\noindent\textbf{Single-step setting.} We compare three single-step scenarios on the PASCAL VOC2007 dataset, where the co-occurrence levels gradually increase in the 19+1, 15+5, and 10+10 settings. As shown in Tab.~\ref{tab: VOC One-step Comparisons}, P$^2$IOD with MS COCO and Objects365 pretrained detectors achieve excellent performance across all experimental settings. Compared to the prompt-pool-based MD-DETR, P$^2$IOD achieves accuracy improvements of 1.6\% / 0.8\%, 2.5\% / 3.9\%, and 8.0\% / 5.2\% in the respective scenarios, indicating that the performance advantage of P$^2$IOD  becomes increasingly significant as the co-occurrence level rises. This trend further demonstrates that P$^2$IOD mitigates the interference caused by prompts pool confusion in co-occurring scenarios. We also compare single-step scenarios on the MS COCO dataset in the Appendix B.1, and the results further demonstrate the effectiveness of P$^2$IOD.

\noindent\textbf{Multi-step setting.}
We compare the multi-step settings on PASCAL VOC2007 and MS COCO datasets. Tab.~\ref{tab: VOC Multi-step Comparisons} shows that on PASCAL VOC2007, the performance degradation of MD-DETR becomes increasingly severe as the number of incremental steps grows. The degradation stems from prompts pool confusion which intensifies as the number of tasks increases. In contrast, our method effectively mitigates such confusion, consistently achieving superior performance across all settings. For the MS COCO dataset, as shown in Tab.~\ref{tab: COCO Multi-step Comparisons}, P$^2$IOD also exhibits the aforementioned advantages. Moreover, P$^2$IOD consistently outperforms other existing approaches across different settings on both datasets, demonstrating its effectiveness and the strong potential of prompt-based techniques in IOD.

\begin{figure*}[th]
    % \vspace{-7mm}
  \centering
  \includegraphics[width=2.0\columnwidth]{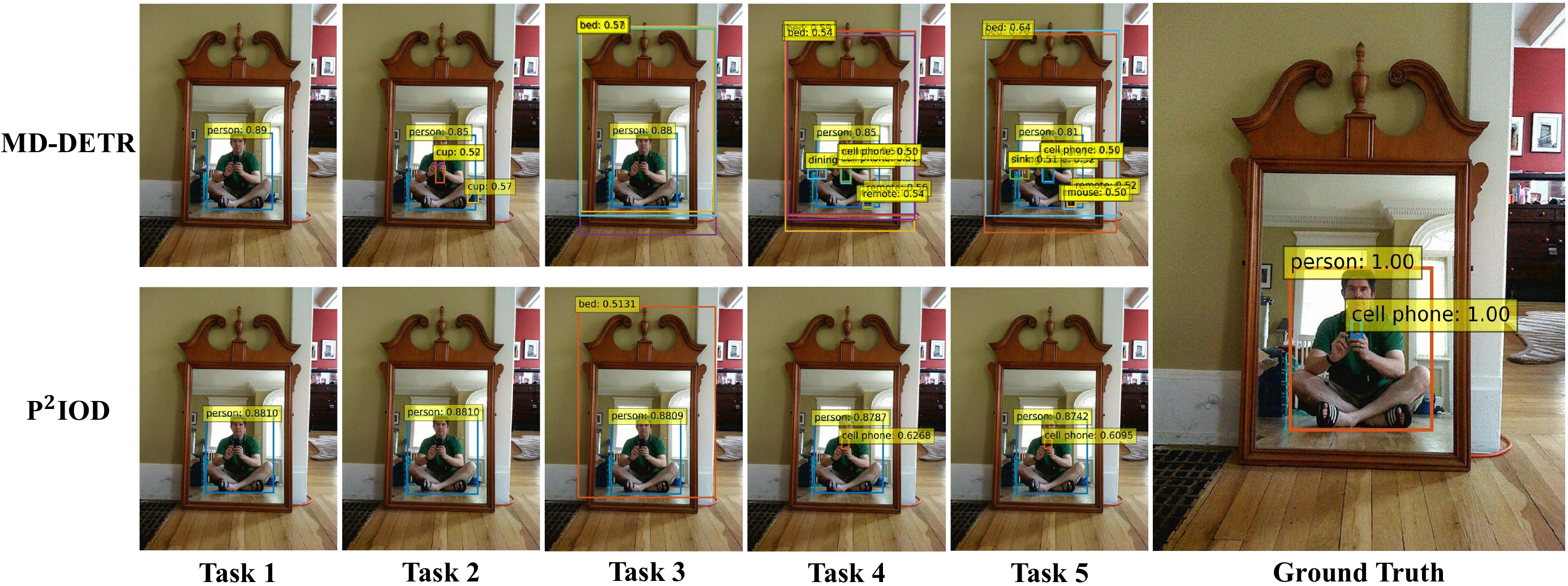}
      \vspace{-2mm}
  \caption{Visualized comparison between P$^2$IOD and MD-DETR. MD-DETR exhibits more false positives and a faster decline in the positive target’s confidence than P$^2$IOD, indicating the impact of prompts pool confusion.}
      \vspace{-4mm}
  \label{fig: Prompt visualize}
\end{figure*}
\subsection{Analysis}

\textbf{Ablation Study.} In Tab.~\ref{tab: ablation study}, categories 1-5 reflect the stability, while categories 6-20 mainly reflect plasticity. After introducing the pseudo-labeling method (b), due to the lack of learnable parameters, pseudo-labeling not only fails to enhance stability but also interferes with current task learning, reducing plasticity. The parameterized prompt structure (c) increases the accuracy of categories 6-20 by 9.5\%, significantly enhancing plasticity, but the accuracy of categories 1-5 drops by 2.6\%, indicating that forgetting still exists. The model fusion (d) alleviates the forgetting problem and balances stability and plasticity to some extent. The sparse loss (f)  reduces parameter conflicts between tasks, improving overall accuracy by 1.1\%. When combined with all methods, it increases the overall accuracy by 9.0\% compared to the baseline. Furthermore, removing the pseudo-labeling (e) results in a significant decrease in stability. The observation demonstrates that our method, by alleviating the prompts pool confusion, allows the pseudo labeling mechanism to effectively mine old-class objects in the background without introducing adverse effects.

\begin{figure}[tb]
    % \vspace{-5mm}
  \centering
  \includegraphics[width=1.0\columnwidth]{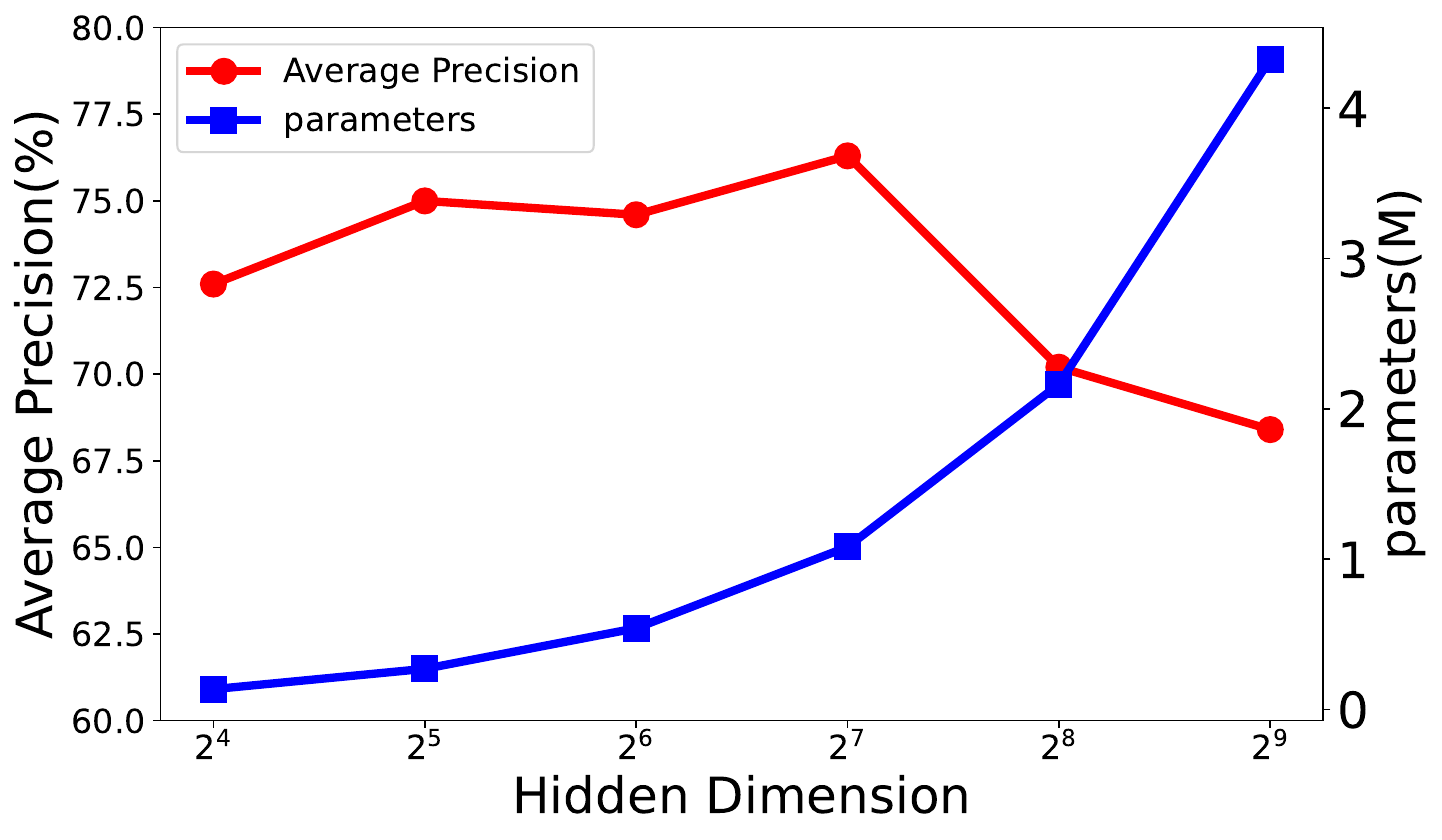}
  \vspace{-7.4mm}
  \caption{Average Precision ($A{P_{50}}$, \%) and parameters (M) on different hidden layer dimensions in the parameterized prompt structure on PASCAL VOC2007 under the 5+5+5+5 setting.}
      \vspace{-5mm}
  \label{fig: hidden_dim}
\end{figure}

\textbf{Impact of Hidden Layer Dimension.} We analyze the impact of the hidden layer dimension in the parameterized prompt structure. The dimension of the hidden layer influences the degree of dimensionality reduction applied to the proposals. We conduct experiments in the PASCAL VOC2007 under the 5+5+5+5 setting. In Fig.~\ref{fig: hidden_dim}, as the hidden layer dimension increases, the model's accuracy initially improves and then declines, suggesting that a moderate increase in the hidden dimension helps retain critical information, while an overly large dimension introduces redundant information that interferes with prompt generation. As the hidden dimension increases, the number of parameters in the parameterized prompt increases accordingly. Our method achieves a favorable trade-off between performance and parameter efficiency (76.3\%, 1.1M) when the hidden dimension is set to 64. In contrast, MD-DETR \cite{bhatt2024preventing} requires more parameters, while simultaneously achieving lower accuracy (61.5\%, 1.8M).

\begin{figure}[tb]
    % \vspace{-5mm}
  \centering
  \includegraphics[width=0.97\columnwidth]{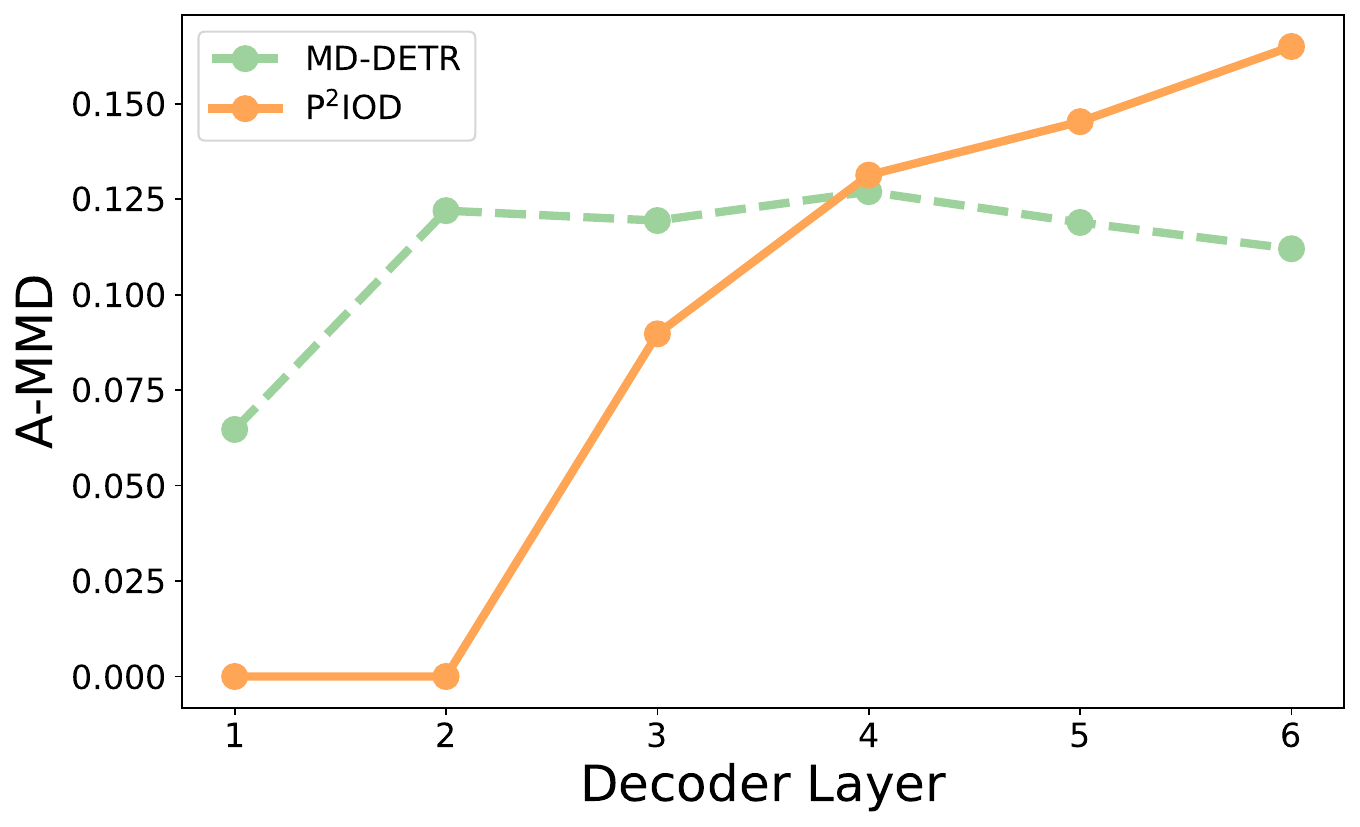}
      \vspace{-4mm}
  \caption{Distribution similarity of prompts across different decoder layers in MD-DETR and P$^2$IOD. A larger A-MMD value indicates a more significant prompt diversity.}
      \vspace{-5mm}
  \label{fig: Prompt Distribution}
\end{figure}

\noindent\textbf{Prompt Comparison.} We compare the distribution similarity of prompts across tasks between P$^2$IOD and MD-DETR \cite{bhatt2024preventing}. We conduct this experiment on PASCAL VOC2007 and use Maximum Mean Discrepancy (MMD) \cite{gretton2012kernel} to evaluate the distribution similarity of prompts, with the average MMD (A-MMD) across all tasks as the evaluation metric. A larger A-MMD value indicates a more significant prompt diversity. As shown in Fig.~\ref{fig: Prompt Distribution}, the diversity of prompts in our P$^2$IOD increases with the depth of decoder layers, and the diversity at the final layer is significantly higher than that of MD-DETR. Our method generates category-independent prompts in shallow layers and category-related prompts in deep layers, aligning well with the multi-layer decoder architecture, while MD-DETR is constrained by its pool structure and struggles to match this characteristic. Furthermore, our method can generate more diverse prompts at the final layer used for object prediction. The variations in prompt distributions highlight the effectiveness of our approach.

\subsection{Visualized Comparison}
We analyze the visualized comparison between P$^2$IOD and MD-DETR to illustrate that the confusion is being addressed. In Fig.~\ref{fig: Prompt visualize}, the visualizations of MD-DETR exhibit numerous false positives, indicating that the confused prompts pool introduces incorrect prompts into the detector, thereby increasing scores for irrelevant categories. In contrast, P$^2$IOD reduces such false positives, demonstrating the effectiveness of our approach in mitigating confusion. We also observe that although P$^2$IOD and MD-DETR have nearly identical confidence in detecting people in the first task, MD-DETR's confidence rapidly declines as the number of tasks increases, suggesting that confusion in MD-DETR undermines confidence in positive objects. In contrast, the confidence in P$^2$IOD remains almost unchanged, proving that our method is unaffected by confusion.

\section{Conclusion}
In this study, we identify a confusion issue within the prompt-pool-based IOD methods. To address this issue, we argue that prompts in IOD should not be exclusively assigned to individual tasks but should instead exhibit adaptive consolidation properties across tasks, with constrained updates. We propose a parameterized prompt structure and parameterized prompt fusion to validate our hypothesis. Experiments on multiple datasets demonstrate that our framework exhibits superior performance. To our knowledge, this is the first work addressing prompts pool confusion in IOD, laying a foundation for broader prompt-based IOD applications.

\section*{Acknowledgments}
\sloppy
This work is partially supported by the Chinese Academy of Sciences 
Project for Young Scientists in Basic Research (YSBR-107).

{
    \small
    \bibliographystyle{ieeenat_fullname}
    \bibliography{main}
}

\clearpage
\setcounter{page}{1}
\appendix
\maketitlesupplementary

\section{Implementation Details}
The pre-trained detectors use the original hyperparameter settings as those in their respective papers \cite{zhu2020deformable, zong2023detrs}. We employ the Adam optimizer with a weight decay of 0.0001. The learning rate is set to 0.0001 for the class embeddings and the parametrized prompt structure, and a lower learning rate of 0.00001 is used for the bounding box embeddings, while freezing the remaining parameters. For training Deformable-DETR \cite{zhu2020deformable} on the PASCAL VOC2007 dataset \cite{everingham2010pascal}, we train each incremental task for 100 epochs, with the learning rate dropped to 0.1 of its original value at the 80th epoch. For training Co-DETR \cite{zong2023detrs} on the PASCAL VOC2007 dataset, we train each incremental task for 12 epochs, with the learning rate dropped to 0.1 of its original value at the 7th epoch. For training Co-DETR on the MS COCO dataset, we train each incremental task for 1 epoch. It is worth noting that in the 19+1 setting on PASCAL VOC2007, the second task contains only 279 images. Due to the limited data, Deformable-DETR tends to overfit during training. To mitigate this issue, we set the hyperparameters of the focal loss to $ \alpha = 0.5 $ and $ \gamma = 3.0 $. The reported accuracy is the average of three independent trials. All experiments are conducted on NVIDIA RTX 4090 GPUs.

\noindent\textbf{Source of Pretrained Models.} We use pretrained detectors available on HuggingFace. Specifically, we employ the Deformable-DETR pretrained on the MS COCO dataset, provided by SenseTime, which can be loaded via the transformers library. Additionally, we use the Co-DETR pretrained on the Objects365 dataset \cite{shao2019objects365}, provided by zongzhuofan, which can be loaded via the mmdetection library.

\begin{algorithm}[t]
\caption{Parameterized Prompts Fusion for Incremental Task $T_t$}
\label{tab: Parameterized Prompts Fusion}
\begin{algorithmic}[1]
\Require $\theta_t$, $\theta_{t-1}^f$, $\theta _{init}$, top-$k\%$, top-$l\%$
\Ensure $\theta_t^f$

\State $\bm{v}_t \gets \theta_t - \theta_{t-1}^f$
\State $\mu_t \gets |\bm{v}_t|$, $\gamma_t \gets \text{sgn}(\bm{v}_t)$
\State $\bm{v}_{t-1}^{f} \gets \theta_{t-1}^f - \theta _{init}$
\State $\mu_{t-1}^{f} \gets |\bm{v}_{t-1}^{f}|$, $\gamma_{t-1}^{f} \gets \text{sgn}(\bm{v}_{t-1}^{f})$

\State $\mathcal{I}_{t-1}^{f} \gets \text{Top-}k\%\text{ indices of } \mu_{t-1}^{f}$
\State $\mathcal{I}_t \gets \text{Top-}l\%\text{ indices of } \mu_t$

\ForAll{parameter index $i$}
    \If{$i \in \mathcal{I}_{t-1}^{f}$}
        \State $\theta_t^f[i] \gets \theta_{t-1}^f[i]$ \Comment{Retain important}
    \ElsIf{$i \in \mathcal{I}_t \setminus \mathcal{I}_{t-1}^{f}$}
        \State $\theta_t^f[i] \gets \theta_t[i]$ \Comment{Retain important}
    \ElsIf{$\gamma_t[i] = \gamma_{t-1}^{f}[i]$}
        \State $\theta_t^f[i] \gets \frac{1}{2} (\theta_t[i] + \theta_{t-1}^f[i])$ \Comment{Average consistent}
    \Else
        \State $\theta_t^f[i] \gets \theta_{t-1}^f[i]$ 
    \EndIf
\EndFor

\State \Return $\theta_t^f$
\end{algorithmic}
\end{algorithm}

\begin{table*}[t]
\centering
\caption{Average precision is compared on the MS COCO dataset under single-step settings of 40+40 and 70+10.}
\label{tab: COCO One-step Comparisons}
\setlength{\tabcolsep}{13.5pt} % 增加列间距
\begin{tabular}{c|c|cccccc}
\hline\hline
Scenarios & Method & $AP$ & $A{P_{50}}$ & $AP_{75}$ & $AP_{S}$ & $AP_{M}$ & $AP_{L}$ \\
\hline
\multirow{6}{*}{40 + 40} & RILOD \cite{li2019rilod} & 29.9 & 45.0 & 32.0 & 15.8 & 33.0 & 40.5 \\
& SID \cite{peng2021sid} & 34.0 & 51.4 & 36.3 & 18.4 & 38.4 & 44.9 \\
& ERD \cite{feng2022overcoming} & 36.9 & 54.5 & 39.6 & 21.3 & 40.4 & 47.5 \\
& CL-DETR \cite{liu2023continual}& 42.0 & 60.1 & 45.9 & 24.0 & 45.3 & 55.6 \\
& LEA \cite{song2025learning}& 41.2 & 59.8 & 44.8 & 25.7 & 45.1 & 54.8 \\
& SDDGR \cite{kim2024sddgr}& 43.0 & 62.1 & 47.1 & 24.9 & 46.9 & 57.0 \\
& GCD \cite{wang2025gcd}& 45.7 & 62.9 & 49.7 & 28.4 & 49.3 & 60.0 \\
& MD-DETR (Objects365)\cite{bhatt2024preventing}& 50.0 & 65.7 & 55.1 & 35.7 & 54.1 & 65.7 \\
& P$^2$IOD (Objects365)& \textbf{54.7} & \textbf{71.1} & \textbf{60.3} & \textbf{39.2} & \textbf{59.4} & \textbf{69.7} \\
\hline
\multirow{6}{*}{70 + 10} & RILOD \cite{li2019rilod} & 24.5 & 37.9 & 25.7 & 14.2 & 27.4 & 33.5 \\
& SID \cite{peng2021sid} & 32.8 & 49.0 & 35.0 & 17.1 & 36.9 & 44.5 \\
& ERD \cite{feng2022overcoming} & 34.9 & 51.9 & 37.4 & 18.7 & 38.8 & 45.5 \\
& CL-DETR \cite{liu2023continual}& 40.4 & 58.0 & 43.9 & 23.8 & 43.6 & 53.5 \\
& LEA \cite{song2025learning}& 45.4 & 65.3 & 49.6 & 28.5 & 48.8 & 59.2 \\
& SDDGR \cite{kim2024sddgr}& 40.9 & 59.5 & 44.8 & 23.9 & 44.7 & 54.0 \\
& GCD \cite{wang2025gcd}& 46.7 & 63.9 & 50.8 & 29.7 & 49.9 & 61.6 \\
& MD-DETR (Objects365)\cite{bhatt2024preventing}& 50.8 & 65.6 & 55.8 & 34.9 & 55.8 & 66.0 \\
& P$^2$IOD (Objects365)& \textbf{55.2} & \textbf{71.9} & \textbf{60.7} & \textbf{41.0} & \textbf{59.7} & \textbf{70.5} \\
\hline\hline
\end{tabular}
\end{table*}

\section{Additional Experiment Results}
\label{app:dwt_details}

\subsection{Single-step Comparison on MS COCO dataset}
We compare two single-step scenarios on the MS COCO dataset, namely the 40+40 and 70+10 settings. As shown in Tab.~\ref{tab: COCO One-step Comparisons}, P$^2$IOD performs excellently across all experimental settings. Compared to the prompt-pool-based MD-DETR, P$^2$IOD achieves $A{P_{50}}$ accuracy improvements of 5.4\% and 6.3\% in the two scenarios, demonstrating that the proposed method effectively addresses the prompts pool confusion. 

% Our method achieves a maximum $A{P_{50}}$ of 71.1\% and 71.9\% in the two settings, outperforming other methods by 8.2\% and 8.0\%. The improvement shows the great potential of prompt-based methods in IOD.

\subsection{Impact of Parameterized Prompt Fusion Threshold}

\begin{table}[t]
\caption{Results ($A{P_{50}}$, \%) on different fusion thresholds in the parameterized prompt fusion on PASCAL VOC2007 under the 10+10 setting.}
\setlength{\tabcolsep}{11.5pt} % 增加列间距
\label{tab: topk}
\begin{tabular}{cc|ccc}
\hline
\multicolumn{2}{c|}{\textbf{Fusion Threshold}}& \multicolumn{3}{c}{\textbf{10+10}} \\
\cline{3-5} 
${\text{top-}}k$&${\text{top-}}l$&\textbf{1-10}&\textbf{11-20}&\textbf{1-20} \\
\hline
\hline
\multicolumn{2}{c|}{no fusion}& 80.72 & 79.54 & 80.13 \\
\hline
0.0 & 0.0 & 81.57 & 79.92 & 80.74 \\
% 0.0 & 0.1 & 81.60 & 80.34 & 80.97 \\
0.0 & 0.3 & 81.16 & 80.06 & 80.61 \\
% 0.0 & 0.5 & 81.57 & 79.70 & 80.64 \\
0.0 & 0.7 & 80.75 & 79.62 & 80.19 \\

% 0.1 & 0.1 & 81.82 & 80.58 & 81.20 \\
% 0.1 & 0.3 & 81.15 & 80.29 & 80.72 \\
% 0.1 & 0.5 & 80.95 & 80.23 & 80.59 \\
% 0.1 & 0.7 & 80.90 & 80.18 & 80.54 \\

% 0.3 & 0.1 & 81.81 & 80.43 & 81.12 \\
0.3 & 0.3 & 81.62 & 80.61 & 81.11 \\
% 0.3 & 0.5 & 81.50 & 80.60 & 81.05 \\
0.3 & 0.7 & 80.46 & 80.55 & 81.00 \\

% 0.5 & 0.1 & 81.85 & 80.32 & 81.08 \\
% 0.5 & 0.3 & 81.87 & 80.51 & 81.19 \\
% 0.5 & 0.5 & 81.88 & 80.47 & 81.17 \\
% 0.5 & 0.7 & 81.87 & 80.49 & 81.18 \\

% 0.7 & 0.1 & 81.73 & 80.31 & 81.02 \\
0.7 & 0.3 & \textbf{82.00} & \textbf{80.42} & \textbf{81.21} \\
% 0.7 & 0.5 & 81.96 & 80.45 & 81.20 \\
0.7 & 0.7 & 81.96 & 80.44 & 81.20 \\

1.0 & - & 81.64 & 79.78 & 80.71 \\
\hline
\end{tabular}
\end{table}

% \begin{table}[t]
% \caption{Results ($A{P_{50}}$, \%) on different fusion thresholds in the parameterized prompt Fusion on PASCAL VOC2007 under the 5+5+5+5 setting.}
% \setlength{\tabcolsep}{3pt} % 增加列间距
% \label{tab: topk}
% \begin{tabular}{cc|ccc}
% \hline
% \multicolumn{2}{c|}{\textbf{Fusion Threshold}}& \multicolumn{3}{c}{\textbf{5+5+5+5}} \\
% \cline{3-5} 
% ${\text{top-}}k$&${\text{top-}}l$&\textbf{1-5 ($T_1$)} & \textbf{6-20 ($T_2+T_3+T_4$)} &\textbf{1-20} \\
% \hline
% \hline
% \multicolumn{2}{c|}{no fusion}& 65.1 & 77.5 & 74.4 \\
% \hline
% 0.0 & 0.0 & 72.5 & 72.6 & 72.5 \\
% 0.0 & 0.3 & 64.8 & 78.2 & 74.8 \\
% 0.3 & 0.0 & 74.9 & 71.0 & 72.0 \\
% 0.3 & 0.3 & 74.0 & 77.2 & 76.4 \\
% % 0.3 & 0.7 & 70.56 & 77.15 & 75.5 \\
% % 0.7 & 0.3 & 74.66 & 72.70 & 73.2 \\
% % 0.7 & 0.7 & 74.26 & 72.93 & 73.3 \\
% 1.0 & - & 74.9 & 68.9 & 70.4 \\
% \hline
% \end{tabular}
% \end{table}

We quantitatively analyze the impact of the ${\text{top-}}k$ and ${\text{top-}}l$ in parameterized prompt fusion. The threshold determines the proportion of the previous and current parameterized prompt structures retained during fusion. To clarify the impact of one-step fusion on accuracy, we perform experiments in the PASCAL VOC2007 under 10+10 setting. As shown in Tab.~\ref{tab: topk}, we observe that neither introducing model fusion (no fusion) nor using only the first task's parameterized prompts (${\text{top-}}k=1.0$) results in poor performance. When ${\text{top-}}k=0.0$ and ${\text{top-}}l=0.0$, the method averages the consistent parameters, improving the performance compared to the non-fused approach, demonstrating that averaging consistent parameters enhances generalization. Furthermore, by comparing different values of ${\text{top-}}k$ and ${\text{top-}}l$, we find that moderately retaining parameters from both the current and previous tasks can further improve performance. The accuracy improvement indicates that retaining key parameters from each task helps preserve task-specific knowledge. We observe the best performance at ${\text{top-}}k=0.7$ and ${\text{top-}}l=0.3$, resulting in a 1.08\% accuracy increase over the non-fused method.

\subsection{Impact of \texorpdfstring{$\lambda$}{lambda} in Sparse Loss}
\begin{table}[t]
\caption{Results ($A{P_{50}}$, \%) on different $\lambda$ in the sparse loss on PASCAL VOC2007 under the 5+5+5+5 setting.}
\label{tab: lambda}
\setlength{\tabcolsep}{8pt} % 增加列间距
\begin{tabular}{c|ccc}
\hline
\multirow{2}{*}{\textbf{$\lambda$}}& \multicolumn{3}{c}{\textbf{5+5+5+5}} \\
\cline{2-4} 
&\textbf{1-5 ($T_1$)} & \textbf{6-20 ($T_2+T_3+T_4$)} &\textbf{1-20} \\
\hline
\hline
0 & 73.1 & 76.0 & 75.3 \\
$1 \times 10^{-6}$ & 73.1 & 76.5 & 75.7 \\
$3 \times 10^{-6}$ & 73.4 & 76.8 & 76.0\\
$1 \times 10^{-5}$ & \textbf{74.0} & \textbf{77.2} & \textbf{76.4} \\
$3 \times 10^{-5}$ & 73.7 & 77.0 & 76.2 \\
$1 \times 10^{-4}$ & 73.9 & 76.8 & 76.1 \\
\hline
\end{tabular}
\end{table}

We quantitatively analyze the impact of $\lambda$, which controls the sparsity of parameterized prompts structure. A larger $\lambda$ enforces sparser weights. We conduct experiments in the PASCAL VOC2007 under the 5+5+5+5 setting. Tab.~\ref{tab: lambda} presents the accuracy results across different $\lambda$ values. We observe that when $\lambda$ is too small (e.g., $1 \times 10^{-6}$), insufficient sparsity fails to concentrate critical knowledge in important parameters, leading to poor performance. Conversely, when $\lambda$ is too large (e.g., $1 \times 10^{-4}$), excessive sparsity limits the capacity for preservation of task knowledge, leading to performance degradation. $\lambda = 1 \times 10^{-5}$ strikes the optimal balance, achieving the highest accuracy of 76.4\% and providing 1.1\% improvement over the baseline without sparse loss. The experiment demonstrates that appropriate sparsity in parameterized prompt structure can help preserve knowledge.

\subsection{Proposal Compression}
MD-DETR \cite{bhatt2024preventing} proposes compressing object-related proposals and introduces a localized query retrieval (QL) method. QL uses a fully connected layer with an input dimension of $N \times D$ and an output dimension of $N$ to generate weights for each proposal. At the same time, QL utilizes the assignment obtained by the Hungarian matching criterion as a supervision signal and constructs a cross-entropy regularization term to supervise the weights of each proposal. However, this design leads to a fully connected layer with a parameter size of $N^2 \times D$, significantly larger than the prompts pool size. The excessive number of parameters not only reduces storage efficiency but also leads to training difficulties. In the code released by MD-DETR, we observe that although the loss in QL does not converge, QL still has a positive effect on accuracy. Inspired by this observation, we designed a Self-Attention Compression (SAC) module to more effectively compress object-related proposals. We first aggregate the feature information of each proposal by applying global average and max pooling operations along the embedding dimension $D$ of the proposal $P$, generating two different proposal descriptors: $F_{avg}$ and $F_{max}$, which represent the average-pooled and max-pooled proposals, respectively. We then apply the shared fully connected layer to both descriptors, fuse the resulting feature vectors through element-wise summation, and apply a sigmoid activation function to obtain the final attention map $M \in \mathbb{R}^{N}$. In short, the attention map is computed as follows:
\begin{equation}
M = \sigma (FC(AvgPool(P)) + FC(MaxPool(P)))
\label{self-attention}
\end{equation}
where $\sigma$ denotes the sigmoid activation function, and $FC$ represents the shared fully connected layer. Compared to the $N^2 \times D$ parameter size required by QL, the fully connected layer in our method contains only $N^2$ parameters, significantly reducing the required number of parameters. To verify the effectiveness of SAC, we first integrate it into MD-DETR \cite{bhatt2024preventing}. We conduct experiments on the PASCAL VOC2007 dataset using the MS COCO pretrained detector. As shown in Tab.~\ref{tab: structrues for proposal compression}, replacing QL with our proposed SAC module improves performance, demonstrating that our method achieves more effective proposal compression with fewer parameters. 

% Additionally, considering that the fully connected layer in the SAC module might cause forgetting problems, we compared a version of the SAC module with no trainable parameters by removing the $FC$ layer. We observed that removing $FC$ layer does not improve stability (1-5), but instead led to a decrease in plasticity (6-20). This indicates that the parameters learned by the $FC$ remain consistent across tasks, thus catastrophic forgetting is not significant.

% \begin{table}[t]
% \caption{Results ($A{P_{50}}$, \%) on different structrues for proposal compression on PASCAL VOC2007 under the 10+10 setting.}
% \setlength{\tabcolsep}{8.5pt} % 增加列间距
% \label{tab: structrues for proposal compression}
% \begin{tabular}{l|ccc}
% \hline
% \multirow{2}{*}{\textbf{Method}}& \multicolumn{3}{c}{\textbf{10+10}} \\
% \cline{2-4} 
% &\textbf{1-10}&\textbf{11-20}&\textbf{1-20} \\
% \hline
% \hline
% MD-DETR - QL & 73.1 & 77.5 & 73.2 \\
% MD-DETR - SAC & 73.6 & 77.7 & 73.6\\
% MD-DETR - SAC w/o $FC$ & 72.8 & 77.3 & 72.9\\
% \hline
% P$^2$IOD - w/o compression & 79.7 & 78.5 & 79.1 \\
% P$^2$IOD - SAC & 79.3 & 77.4 & 78.3 \\
% P$^2$IOD - SAC w/o $FC$ & 77.1 & 73.2 & 75.2 \\
% P$^2$IOD - average & \textbf{82.0} & \textbf{80.4} & \textbf{81.2} \\
% \hline
% \end{tabular}
% \end{table}

\begin{table}[t]
\caption{Results ($A{P_{50}}$, \%) on different structrues for proposal compression on PASCAL VOC2007 under the 5+5+5+5 setting.}
\setlength{\tabcolsep}{8.5pt} % 增加列间距
\label{tab: structrues for proposal compression}
\begin{tabular}{l|ccc}
\hline
\multirow{2}{*}{\textbf{Method}}& \multicolumn{3}{c}{\textbf{5+5+5+5}} \\
\cline{2-4} 
&\textbf{1-5}&\textbf{6-20}&\textbf{1-20} \\
\hline
\hline
MD-DETR - QL & 55.2 & 63.6 & 61.5 \\
MD-DETR - SAC & 68.9 & 62.3 & 64.0\\
% MD-DETR - SAC w/o $FC$ & 60.8 & 59.6 & 59.9\\
MD-DETR - average & 39.0 & 45.0 & 43.5\\
\hline
P$^2$IOD - SAC & 40.9 & 72.6 & 64.7 \\
% P$^2$IOD - SAC w/o $FC$ & 45.3 & 69.9 & 63.7 \\
P$^2$IOD - w/o compression & 45.9 & 70.5 & 64.3 \\
P$^2$IOD - average & \textbf{74.0} & \textbf{77.2} & \textbf{76.4} \\
\hline
\end{tabular}
\end{table}

\begin{table}[t]
\caption{Lookup table for variable definition in the paper.}
\label{tab: variable definitions}
\begin{tabular}{cc}
\hline
Variable & Definition\\
\hline
${T_t}$ & task $t$\\
$\theta ^ * $ & frozen parameters \\
$\theta $ & parametrized prompt\\
$\theta _t$ & parameterized prompt after task $t$ training\\
$\theta _t^f$ & parameterized prompt after task $t$ fusion\\
${\bm{v}_t}$ & task vector\\
$\mu _t$ & magnitude of task vector\\
$\mathcal{I}_t$ & top indices of $\mu _t$\\
$\gamma _t$ & direction of task vector\\
$x$ & input image\\
$P$ & proposals\\
$N$ & number of proposals\\
$D$ & dimension of embedding\\
$Q$ & query function\\
$p$ & prompts\\
${L_p}$ & length of prompts\\
$d$ & hidden layer dimension\\
$W$ & weight of MLP layer\\
${L_s}$ & sparse loss\\
$\lambda$ & sparse loss hyperparameter\\
${{\hat y}_i}$ & detector prediction\\
${{\hat s}_i}$ & score for prediction\\
${{\hat b}_i}$ & bounding box coordinates for prediction\\
${{\tilde y}_i}$ & pseudo label\\
${{\tilde c}_i}$ & category for pseudo label\\
${{\tilde b}_i}$ & bounding box coordinates for pseudo label\\
$ \tau $ & threshold of Pseudo Labeling\\
\hline
\end{tabular}
\end{table}

We integrate the proposed SAC module into P$^2$IOD, and an inconsistency arises compared to its integration into MD-DETR. As shown in Tab.~\ref{tab: structrues for proposal compression}, adding the SAC module does not improve accuracy as in MD-DETR but instead reduces performance. Averaging all background and object proposals achieves the best performance. We attribute this discrepancy to differences in prompt structure. In MD-DETR, the prompt matching mechanism in the prompts pool requires query features to be task-discriminative, so removing background information enables better matching. Additionally, the prompts pool stores a limited number of prompts and linearly combines them. The limited representational capacity forces the prompts pool to retain only the most distinctive information, namely, object-related information. In contrast, the parameterized prompts in P$^2$IOD act as a mapping from the proposal domain to the prompt domain, offering a larger representational space that allows for the inclusion of background information. In detection tasks, including background information helps the detector distinguish between foreground and background. Therefore, removing background information via the SAC module results in suboptimal prompts. The above experiments suggest that in prompt-based IOD, background-related knowledge should be incorporated into the prompts. The prompts pool is not suitable for IOD due to its structural limitations in providing background knowledge. Furthermore, as shown in Tab.~\ref{tab: structrues for proposal compression}, removing the averaging operation leads to performance degradation, indicating that proposal compression is necessary. Without proposal compression, the number of parameters in the parameterized prompts increases dramatically. Although this setup preserves complete proposal information, the excessive number of trainable parameters hinders its convergence. The averaging operation maintains a reasonable parameter size while simultaneously retaining both background and object information. Therefore, P$^2$IOD adopts averaging as its proposal compression strategy.

% We integrate the proposed SAC module into P$^2$IOD. As shown in Tab.~\ref{tab: structrues for proposal compression}, adding SAC module does not improve accuracy as in MD-DETR but instead decreases performance. We attribute this inconsistency to differences in prompt structure. In MD-DETR, the prompts pool stores a limited number of prompts and linearly combines the stored prompts. The limited capacity of the prompts pool forces it to retain only the most distinctive information, that is, object-related information. Therefore, removing background information by the proposal compression can more effectively match the prompts. In contrast, the parameterized prompts in P$^2$IOD act as a mapping from the proposal domain to the prompt domain with a larger representational space, allowing background information to be included. Removing background information through SAC module therefore leads to suboptimal prompts. We also observe that removing proposal compression degrades performance. Without compression, the number of parameters in the parameterized prompts increases dramatically. Although this setting preserves complete proposal information, the excessive number of trainable parameters hinders convergence. In contrast, simple averaging achieves higher accuracy while maintaining a reasonable parameter size. Therefore, P$^2$IOD adopts averaging as its proposal compression strategy.

\section{More Explanations}
This section provides more details about the parameterized prompt fusion algorithm and variable definition.

\subsection{Parameterized Prompt Fusion}
The details of the parameterized prompt fusion algorithm are presented in Alg.~\ref{tab: Parameterized Prompts Fusion}.

\subsection{Variable Definitions}
All variable definitions used in our method are listed in Tab.~\ref{tab: variable definitions}.

\end{document}